\pdfoutput=1
\documentclass[10pt,twocolumn,letterpaper]{article}

\usepackage{cvpr}
\usepackage{times}
\usepackage{epsfig}
\usepackage{graphicx}
\usepackage{amsmath}
\usepackage{amssymb}
\usepackage{tabularx}

\usepackage[usenames,dvipsnames,svgnames,table]{xcolor}

\usepackage{comment}
\usepackage{longtable}
\usepackage[latin1]{inputenc}
\usepackage{latexsym}
\usepackage{amsthm}
\usepackage{enumerate}
\usepackage{amsfonts}
\usepackage[format=plain, font=small, labelfont=bf]{caption}
\usepackage{fancyhdr}
\usepackage{lscape}
\usepackage[T1]{fontenc}
\usepackage{floatrow}
\usepackage{enumerate}
\usepackage{xcolor}
\usepackage{subcaption}
\usepackage{caption}
\usepackage{soul}
\usepackage{booktabs}

\theoremstyle{remark}

\theoremstyle{definition}
\newtheorem{definition}{Definition}

\RequirePackage{color}
\definecolor{RED}{rgb}{1,0,0}
\definecolor{BLUE}{rgb}{0,0,1}
\definecolor{White}{rgb}{1,1,1}

\definecolor{celadon}{rgb}{0.67, 0.88, 0.69}

\newcommand{\N}{{\mathcal{N}}}
\newcommand{\R}{{\mathbb{R}}}

\newcommand{\Expect}{{\mathbb{E}}}
\newcommand{\iver}[1]{{\mathbb{[} #1 \mathbb{]}}}

\newcommand{\norm}[1]{\left\lVert#1\right\rVert}

\usepackage[pagebackref=true,breaklinks=true,letterpaper=true,colorlinks,bookmarks=false]{hyperref}

\cvprfinalcopy 

\def\cvprPaperID{2479} 
\def\httilde{\mbox{\tt\raisebox{-.5ex}{\symbol{126}}}}

\ifcvprfinal\fi
\begin{document}

\title{Semi-Supervised Learning with the Deep Rendering Mixture Model}

\author{Tan Nguyen$^{1,2}$ \qquad  Wanjia Liu$^{1}$ \qquad  Ethan Perez$^{1}$ \qquad Richard G. Baraniuk$^{1}$ \qquad Ankit B. Patel$^{1,2}$\\
$^{1}$Rice University \qquad $^{2}$Baylor College of Medicine\\
6100 Main Street, Houston, TX 77005 \qquad 1 Baylor Plaza, Houston, TX 77030\\
{\tt\small \{mn15, wl22, ethanperez, richb\}@rice.edu} \qquad {\tt\small ankitp@bcm.edu}
}

\newcommand {\note}[1]{{\color{blue}\sf{[#1]}}}
\newcommand {\richb}[1]{{\color{green}\sf{[RB: #1]}}}
\newcommand {\tnote}[1]{{\color{blue}\sf{[TN: #1]}}}
\newcommand {\abp}[1]{{\color{purple}\sf{[AP: #1]}}}

\maketitle

\begin{figure*}[t!]
       \centering
 	 \includegraphics[width=1.0\textwidth]{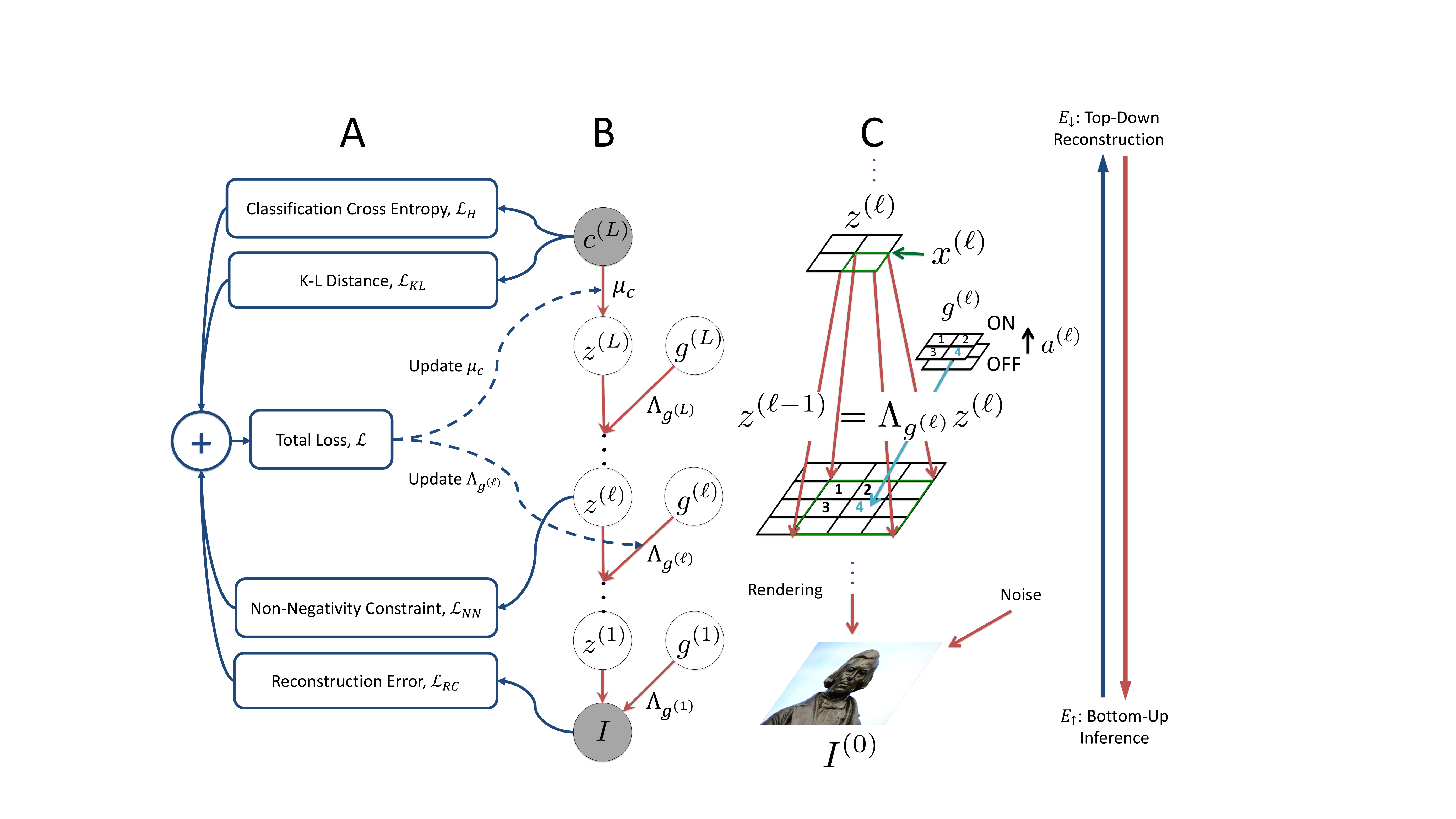}
	 \caption{Semi-supervised learning for Deep Rendering Mixture Model (DRMM) (A) Computation flow for semi-supervised DRMM loss function and its components. Dashed arrows indicate parameter update. (B) The Deep Rendering Mixture Model (DRMM). All dependence on pixel location $x$ has been suppressed for clarity. (C) DRMM generative model: a single super pixel $x^{(\ell)}$ at level $\ell$ (green, upper) renders down to a $3\times 3$ image patch at level $\ell-1$ (green, lower), whose location is specified by $g^{(\ell)}$ (light blue). (C) shows only the transformation from level $\ell$ of the hierarchy of abstraction to level $\ell-1$.}
	 \label{fig:overall-model}
\end{figure*} 
\begin{abstract}
  
   Semi-supervised learning algorithms reduce the high cost of acquiring labeled training data by using both labeled and unlabeled data during learning. Deep Convolutional Networks (DCNs) have achieved great success in supervised tasks and as such have been widely employed in the semi-supervised learning. In this paper we leverage the recently developed Deep Rendering Mixture Model (DRMM), a probabilistic generative model that models latent nuisance variation, and whose inference algorithm yields DCNs. We develop an EM algorithm for the DRMM to learn from both labeled and unlabeled data. Guided by the theory of the DRMM, we introduce a novel non-negativity constraint and a variational inference term. We report state-of-the-art performance on MNIST and SVHN and competitive results on CIFAR10. We also probe deeper into how a DRMM trained in a semi-supervised setting represents latent nuisance variation using synthetically rendered images. Taken together, our work provides a unified framework for supervised, unsupervised, and semi-supervised learning.

\end{abstract}

\section{Introduction}

Humans are able to learn from both labeled and unlabeled data. Young infants can acquire knowledge about the world and distinguish objects of different classes with only a few provided ``labels''. Mathematically, this poverty of input implies that the data distribution $p(I)$ contains information useful for inferring the category posterior $p(c|I)$. The ability to extract this useful hidden knowledge from the data in order to leverage both labeled and unlabeled examples for inference and learning, i.e. semi-supervised learning, has been a long-sought after objective in computer vision, machine learning and computational neuroscience. 

In the last few years, Deep Convolutional Networks (DCNs) have emerged as powerful supervised learning models that achieve near-human or super-human performance in various visual inference tasks, such as object recognition and image segmentation. However, DCNs are still far behind humans in semi-supervised learning tasks, in which only a few labels are available. The main difficulty in semi-supervised learning in DCNs is that, until recently, there has not been a mathematical framework for deep learning architectures. As a result, it is not clear how DCNs encode the data distribution, making combining supervised and unsupervised learning challenging.

Recently, the Deep Rendering Mixture Model (DRMM) \cite{patel2015probabilistic, patel2016probabilistic} has been developed as a probabilistic graphical model underlying DCNs. The DRMM is a hierarchical generative model in which the image is rendered via multiple levels of abstraction. It has been shown that the bottom-up inference in the DRMM corresponds to the feedforward propagation in the DCNs. The DRMM enables us to perform semi-supervised learning with DCNs. Some preliminary results for semi-supervised learning with the DRMM are provided in \cite{patel2016probabilistic}. Those results are promising, but more work is needed to evaluate the algorithms across many tasks.

In this paper, we systematically develop a semi-supervised learning algorithm for the Non-negative DRMM (NN-DRMM), a DRMM in which the intermediate rendered templates are non-negative. Our algorithm contains a bottom-up inference pass to infer the nuisance variables in the data and a top-down pass that performs reconstruction. We also employ variational inference and the non-negative nature of the NN-DRMM to derive two new penalty terms for the training objective function. An overview of our algorithm is given in Figure~\ref{fig:overall-model}. We validate our methods by showing state-of-the-art semi-supervised learning results on MNIST and SVHN, as well as comparable results to other state-of-the-art methods on CIFAR10. Finally, we analyze the trained model using a synthetically rendered dataset, which mimics CIFAR10 but has ground-truth labels for nuisance variables, including the orientation and location of the object in the image. We show how the trained NN-DRMM encodes nusiance variations across its layers and show a comparison against traditional DCNs.

\section{Related Work}
We focus our review on semi-supervised methods that employ neural network structures and divide them into different types.

\noindent \textbf{Autoencoder-based Architectures:} Many early works in semi-supevised learning for neural networks are built upon autoencoders \cite{bengio2009learning}. In autoencoders, the images are first projected onto a low-dimensional manifold via an encoding neural network and then reconstructed using a decoding neural network. The model is learned by minimizing the reconstruction error. This method is able to learn from unlabeled data and can be combined with traditional neural networks to perform semi-supervised learning. In this line of work are the Contractive Autoencoder \cite{rifai2011contractive}, the Manifold Tangent Classifier, the Pseudo-label Denoising Auto Encoder \cite{lee2013pseudo}, the Winner-Take-All Autoencoders \cite{makhzani2015winner}, and the Stacked What-Where Autoencoder \cite{zhao2015swwae}. These architectures perform well when there are enough labels in the dataset but fail when the number of labels is reduced since the data distribution is not taken into account. Recently, the Ladder Network \cite{rasmus2015semi} was developed to overcome this shortcoming. The Ladder Network approximates a deep factor analyzer where each layer in the model is a factor analyzer. Deep neural networks are then used to do approximate bottom-up and top-down inference.\\

\noindent \textbf{Deep Generative Models:} Another line of work in semi-supervised learning is to use neural networks to estimate the parameters of a probabilistic graphical model. This approach is applied when the inference in the graphical model is hard to derive or when the exact inference is computationally intractable. The Deep Generative Model family  is in this line of work \cite{kingma2014semi, maaloe2016auxiliary}.\\ 

Both Ladder Networks and Deep Generative Models yield good semi-supervised learning performance on benchmarks. They are complementary to our semi-supervised learning on DRMM. However, our method is different from these approaches in that the DRMM is the graphical model underlying DCNs, and we theoretically derive our semi-supervised learning algorithm as a proper probabilistic inference against this graphical model. \\ 

\noindent \textbf{Generative Adversarial Networks (GANs):} In the last few years a variety of GANs have achieved promising results in semi-supervised learning on different benchmarks, including MNIST, SVHN and CIFAR10. These models also generate good-looking images of natural objects. In GANs, two neural networks play a minimax game. One generates images, and the other classifies images. The objective function is the game's Nash equilibrium, which is different from standard object functions in probabilistic modeling. It would be both exciting and promising to extend the DRMM objective to a minimax game as in GANs, but we leave this for future work.  

\section{Deep Rendering Mixture Model}
\label{sec:drmm}
The Deep Rendering Mixture Model (DRMM) is a recently developed probabilistic generative model whose bottom-up inference, under a non-negativity assumption, is equivalent to the feedforward propagation in a DCN \cite{patel2015probabilistic, patel2016probabilistic}. It has been shown that the inference process in the DRMM is efficient due to the hierarchical structure of the model. In particular, the latent variations in the data are captured across multiple levels in the DRMM. This factorized structure results in an exponential reduction of the free parameters in the model and enables efficient learning and inference. The DRMM can potentially be used for semi-supervised learning tasks \cite{patel2015probabilistic}.

\begin{definition}[\textbf{Deep Rendering Mixture Model}] \label{defn:drmm}
The \textit{Deep Rendering Mixture Model (DRMM)} is a deep Gaussian Mixture Model (GMM) with special constraints on the latent variables. 
Generation in the DRMM takes the form:
\begin{align}
    c^{(L)} &\sim \textrm{Cat}(\{\pi_{c^{(L)}}\}), \quad
    g^{(\ell)} \sim \textrm{Cat}(\{\pi_{g^{(\ell)}}\})
\label{eq:gen1}\\
    \mu_{cg} &\equiv \Lambda_{g}\mu_{c^{(L)}}
             \equiv \Lambda^{(1)}_{g^{(1)}}\Lambda^{(2)}_{g^{(2)}} \dots \Lambda^{(L-1)}_{g^{(L-1)}}\Lambda^{(L)}_{g^{(L)}} \mu_{c^{(L)}} \label{eqn:drmm} 
             \\
    I &\sim \N(\mu_{cg},\,\sigma^{2}\textbf{1}_{D^{(0)}}),
    \label{eq:gen3}
\end{align}
where $\ell\in [L] \equiv \{1,2,\dots,L\}$ is the layer, $c^{(L)}$ is the object category, $g^{(\ell)}$ are the latent (nuisance) variables at layer $\ell$, and $\Lambda_{g^{(\ell)}}^{(\ell)}$ are parameter dictionaries that contain templates at layer $\ell$. Here the image $I$ is generated by adding isotropic Gaussian noise to a multiscale ``rendered'' template $\mu_{cg}$.
\end{definition}

In the DRMM, the \emph{rendering path} is defined as the sequence $(c^{(L)},g^{(L)},\ldots,g^{(\ell)},\ldots,g^{(1)})$ from the root (overall class) down to the individual pixels at $\ell=0$. 
The variable $\mu_{cg}$ is the template used to render the image, and $\prod_{\ell} \Lambda_{g^{(\ell)}}^{(L)}\mu_{c^{(L)}}$ represents the sequence of local nuisance transformations that partially render finer-scale details as we move from abstract to concrete. Note that the factorized structure of the DRMM results in an {\it exponential reduction in the number of free parameters}. This enables efficient inference, learning, and better generalization.\\

A useful variant of the DRMM is the Non-Negative Deep Rendering Mixture Model (NN-DRMM), where the intermediate rendered templates are constrained to be non-negative. The NN-DRMM model can be written as
\begin{align}
     z^{(\ell)}_{n} &= \Lambda_{g^{(\ell + 1)}_{n}}^{(\ell + 1)} \cdots \Lambda_{g^{(L)}_{n}}^{(L)} \mu_{c^{(L)}_{n}} \ge 0 \quad \forall \ell \in \{1,\dots,L\}.
\end{align}


It has been proven that the inference in the NN-DRMM via a dynamic programming algorithm leads to the feedforward propagation in a DCN. This paper develops a semi-supevised learning algorithm for the NN-DRMM. For brevity, throughout the rest of the paper we will drop the NN. \\

\begin{figure*}[t!]
    \centering
        \includegraphics[width=1.0\textwidth]{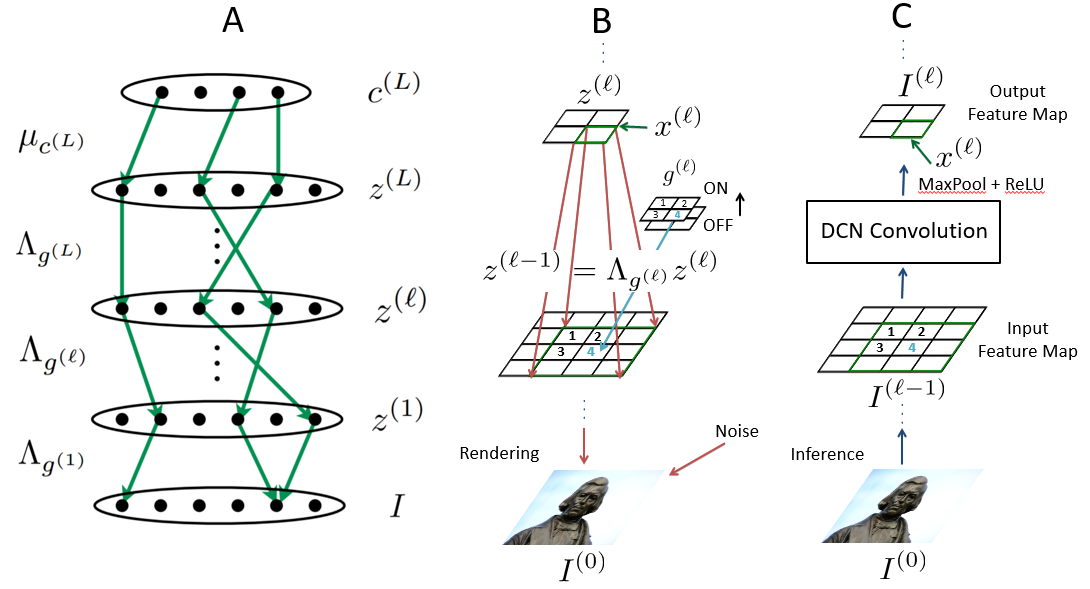}
        \caption{(A)The Sum-over-Paths Formulation of the DRMM. Each rendering path contributes only if it is active (green)\cite{patel2016probabilistic}. While exponentially many possible rendering paths exist, only a very small fraction are active. (B) Rendering from layer $\ell \rightarrow \ell-1$ in the DRMM. (C) Inference in the Nonnegative DRMM leads to processing identical to the DCN.}
        \label{fig:deep-sparse-coding}
\end{figure*}

\noindent\textbf{Sum-Over-Paths Formulation of the DRMM:} The DRMM can be can be reformulated by expanding out the matrix multiplications in the generation process into scalar products. Then each pixel intensity $I_x = \sum_{p} \lambda_p^{(L)} a_p^{(L)} \cdots \lambda_p^{(1)} a_p^{(1)}$ is the sum over all \textit{active paths} leading to that pixel of the product of weights along that path. The sparsity of $a$ controls the number fraction of active paths. Figure~\ref{fig:deep-sparse-coding} depicts the sum-over-paths formulation graphically.

\section{DRMM-based Semi-Supervised Learning}

\subsection{Learning Algorithm}
\label{sec:semi-sup algo}

 Our semi-supervised learning algorithm for the DRMM is analogous to the hard Expectation-Maximization (EM) algorithm for GMMs \cite{bishop2006pattern, patel2015probabilistic, patel2016probabilistic}. In the E-step, we perform a bottom-up inference to estimate the most likely joint configuration of the latent variables $\hat{g}$ and the object category $\hat{c}$ given the input image. This bottom-up pass is then followed by a top-down inference which uses $(\hat{c},\hat{g})$ to reconstruct the image $\hat{I} \equiv \mu_{\hat{c} \hat{g}}$ and compute the reconstruction error $\mathcal{L}_{RC}$. It is known that when applying a hard EM algorithm on GMMs, the reconstruction error averaged over the dataset is proportional to the expected complete-data log-likelihood. For labeled images, we also compute the cross-entropy $\mathcal{L}_{H}$ between the predicted object classes and the given labels as in regular supervised learning tasks. In order to further improve the performance, we introduce a Kullback-Leibler divergence penalty $\mathcal{L}_{KL}$ on the predicted object class $\hat{c}$ and a non-negativity penalty $\mathcal{L}_{NN}$ on the intermediate rendered templates $z^{(\ell)}_{n}$ at each layer into the training cost objective function. The motivation and derivation for these two terms are discussed in section \ref{sec:var-inf} and \ref{sec:nn-opt} below. The final objective function for semi-supervised learning in the DRMM is given by $\mathcal{L} \equiv \alpha_{H}\mathcal{L}_{H} + \alpha_{RC}\mathcal{L}_{RC} + \alpha_{KL}\mathcal{L}_{KL} + \alpha_{NN}\mathcal{L}_{NN}$, where $\alpha_{CE}$, $\alpha_{RC}$, $\alpha_{KL}$ and $\alpha_{NN}$ are the weights for the cross-entropy loss $\mathcal{L}_{CE}$, reconstruction loss $\mathcal{L}_{RC}$, variational inference loss $\mathcal{L}_{KL}$, and the non-negativity penalty loss $\mathcal{L}_{NN}$, respectively. The losses are defined as follows:
\begin{align} 
	\label{eqn:semisup-cost}
	\mathcal{L}_{H} &\equiv -\frac{1}{|\mathcal{D}_{L}|}\sum_{n\in \mathcal{D}_{L}}\sum_{c\in\mathcal{C}} \iver{\hat{c}_{n}=c_{n}}\log q\left(c|I_{n}\right) \\
	\mathcal{L}_{RC} &\equiv \frac{1}{N}\sum_{n=1}^{N}\norm{I_{n} - \hat{I}_{n}}_{2}^{2} \\
	\mathcal{L}_{KL} &\equiv \frac{1}{N}\sum_{n=1}^{N}\sum_{c\in\mathcal{C}}q\left(c|I_{n}\right)\log\left(\frac{q\left(c|I_{n}\right)}{p(c)}\right) \\
	\mathcal{L}_{NN} &\equiv \frac{1}{N}\sum_{n=1}^{N}\sum_{\ell=1}^{L} \norm{\max\left\{0, -z_{n}^{\ell}\right\}}_{2}^{2}.
\end{align}

Here, $q\left(c|I_{n}\right)$ is an approximation of the true posterior $p(c|I_{n})$. In the context of the DRMM and the DCN, $q\left(c|I_{n}\right)$ is the SoftMax activations, $p(c)$ is the class prior, $\mathcal{D}_{L}$ is the set of labeled images, and $\iver{\hat{c}_{n}=c_{n}} = 1$ if $\hat{c}_{n}=c_{n}$ and $0$ otherwise. The $\max\{0,\cdot\}$ operator is applied element-wise and equivalent to the ReLu activation function used in DCNs.

During training, instead of a closed-form M step as in EM algorithm for GMMs, we use gradient-based optimization methods such as stochastic gradient descent to optimize the objective function. 
\subsection{Variational Inference}
\label{sec:var-inf}
The DRMM can compute the most likely latent configuration $(\hat{c}_n, \hat{g}_{n})$ given the image $I_{n}$, and therefore, allows the exact inference of $p(c, \hat{g}_{n}|I_{n})$. Using variational inference, we would like the approximate posterior $q(c, \hat{g}_{n}|I_{n}) \equiv\underset{g \in \mathcal{G}}{\max} p(c,g|I_{n})$  to be close to the true posterior $p(c|I)=\underset{g\in\mathcal{G}}{\sum}p(c,g|I_{n})$ by minimizing the KL divergence $\mathcal{D}\left(q(c, \hat{g}_{n}|I_{n})\|p(c|I_{n})\right)$. It has been shown in \cite{blei2016variational} that this optimization is equivalent to the following:
\begin{align}
    \min_{q} -\Expect_q \left[ \ln p(I_{n}|z) \right] + D_{KL}(q(z|I_{n}) || p(z)),
    \label{eqn:var-inf}
\end{align}
where $z=c$. A similar idea has been employed in variational autoencoders \cite{kingma2014adam}, but here instead of using a Gaussian distribution, $z$ is a categorical random variable. An extension of the optimization~\ref{eqn:var-inf} is given by:
\begin{align}
    \min_{q} -\Expect_q \left[ \ln p(I_{n}|z) \right] + \beta D_{KL}(q(z|I_{n}) || p(z)).
    \label{eqn:irina-inf}
\end{align}
As has been shown in \cite{higgins2016early}, for this optimization, there exists a value for $\beta$ such that latent variations in the data are optimally disentangled. 

The KL divergence in Eqns.~\ref{eqn:var-inf} and \ref{eqn:irina-inf} results in the $\mathcal{L}_{KL}$ loss in the semi-supervised learning objective function for DRMM (see Eqn.~\ref{eqn:semisup-cost}). Similarly, the expected reconstruction error $-\Expect_q \left[ \ln p(I_{n}|z) \right]$ corresponds to the $\mathcal{L}_{RC}$ loss in the objective function. Note that this expected reconstruction error can be exactly computed for the DRMM since there are only a finite number of configurations for the class $c$. When the number of object classes is large, such as in ImageNet \cite{russakovsky2015imagenet} where there are 1000 classes, sampling techniques can be used to approximate  $-\Expect_q \left[ \ln p(I_{n}|z) \right]$. From our experiments (see Section~\ref{sec:experiments}), we notice that for semi-supervised learning tasks on MNIST, SVHN, and CIFAR10, using the most likely $\hat{c}$ predicted in the bottom-up inference to compute the reconstruction error yields the best classification accuracies. 
\subsection{Non-Negativity Constraint Optimization}
\label{sec:nn-opt}

In order to derive the DCNs from the DRMM, the intermediate rendered templates $z_{n}^{\ell}$ must be non-negative\cite{patel2016probabilistic}. This is necessary in order to apply the max-product algorithm, wherein we can push the max to the right to get an efficient message passing algorithm. We enforce this condition in the Top-Down inference of the DRMM by introducing new non-negativity constrains $z_{n}^{\ell} \ge 0 \,\,,\forall \ell \in \{1,\dots,L\} $ into the optimization \ref{eqn:var-inf} and \ref{eqn:irina-inf}. There are various well-developed methods to solve optimization problems with non-negativity constraints. We employ a simple but useful approach, which adds an extra non-negativity penalty, in this case, $\frac{1}{N}\sum_{n=1}^{N}\sum_{\ell=1}^{L} \norm{\max\left\{0, -z_{n}^{\ell}\right\}}_{2}^{2}$, into the objective function.  This yields an unconstrained optimization which can be solved by gradient-based methods such as stochastic gradient descent. We cross-validate the penalty weight $\alpha_{NN}$.
\section{Experiments}
\label{sec:experiments}
We evaluate our methods on the MNIST, SVHN, and CIFAR10 datasets. In all experiments, we perform semi-supervised learning using the DRMM with the training objective including the cross-entropy cost, the reconstruction cost, the KL-distance, and the non-negativity penalty discussed in Section~\ref{sec:semi-sup algo}. We train the model on all provided training examples with different numbers of labels and report state-of-the-art test errors on MNIST and SVHN. The results on CIFAR10 are comparable to state-of-the-art methods. In order to focus on and better understand the impact of the KL and NN penalties on the semi-supervised learning in the DRMM, we don't use any other regularization techniques such as DropOut or noise injection in our experiments. We also only use a simple stochastic gradient descent optimization with exponentially-decayed learning rates to train the model. Applying regularization and using better optimization methods like ADAM \cite{kingma2014adam} may help improve the semi-supervised learning performance of the DRMM. More model and training details are provided in the Appendix.

\subsection{MNIST}
MNIST dataset contains 60,000 training images and 10,000 test images of handwritten digits from 0 to 9. Each image is of size 28-by-28. For evaluating semi-supervised learning, we randomly choose $N_{L}\in \{50, 100, 1000\}$ images with labels from the training set such that the amounts of labeled training images from each class are balanced. The remaining training images are provided without labels. We use a 5-layer DRMM with the feedforward configuration similar to the Conv Small network in \cite{rasmus2015semi}. We apply batch normalization on the net inputs and use stochastic gradient descent with exponentially-decayed learning rate to train the model. 

Table \ref{tab:test-error-mnist} shows the test error for each experiment. The KL and NN penalties help improve the semi-supervised learning performance across all setups. In particular, the KL penalty alone reduces the test error from $13.41\%$ to $1.36\%$ when $N_{L}=100$. Using both the KL and NN penalties, the test error is reduced to $0.57\%$, and the DRMM achieves state-of-the-art results in all experiments. \footnote{The results for improved GAN is on permutation invariant MNIST task while the DRMM performance is on the regular MNIST task. Since the DRMM contains local latent variables $t$ and $a$ at each level, it is not suitable for tasks such as permutation invariant MNIST} We also analyze the value that the KL and NN penalties add to the learning. Table~\ref{tab:value-added-analysis} reports the reductions in test errors for $N_{L}\in\{50, 100, 1K\}$ when using the KL penalty only, the NN penalty only, and both penalties during training. Individually, the KL penalty leads to significant improvements in test errors ($12.05\%$ reduction in test error when $N_{L}=100$), likely since it helps disentangle latent variations in the data. In fact, for a model with continuous latent variables, it has been experimentally shown that there exists an optimal value for the KL penalty $\alpha_{KL}$ such that all of the latent variations in the data are almost optimally disentangled   \cite{higgins2016early}. More results are provided in the Appendix. 

\begin{table*}[!t]
\vspace{1mm}
\centering
\small
\begin{tabular}{@{} cccc @{}}
\hline
\multicolumn{1}{c}{Model} &\multicolumn{3}{c}{Test error (\%) for a given number of labeled examples} \\
\multicolumn{1}{c}{\bf } &\multicolumn{1}{c}{\bf $N_{L}=50$} &\multicolumn{1}{c}{\bf $N_{L}=100$} &\multicolumn{1}{c}{\bf $N_{L}=1K$}
\\ \hline \hline
DGN \cite{kingma2014semi} & - & $3.33\pm0.14$ & $2.40\pm0.02$\\
catGAN \cite{springenberg2015unsupervised} & - & $1.39\pm0.28$ & - \\
Virtual Adversarial \cite{miyato2015distributional} & - & $2.12$ & - \\
Skip Deep  Generative Model \cite{maaloe2016auxiliary} & - & $1.32$ & - \\
LadderNetwork \cite{rasmus2015semi} & - & $1.06\pm0.37$ & $0.84\pm0.08$ \\
Auxiliary Deep  Generative Model \cite{maaloe2016auxiliary} & - & $0.96$ & - \\
ImprovedGAN \cite{salimans2016improved} & $2.21\pm1.36$ & $0.93 \pm0.065$ & - \\
\hline
DRMM 5-layer & $21.73$ & $13.41$ & $2.35$ \\
DRMM 5-layer + NN penalty & $22.10$ & $12.28$ & $2.26$ \\
DRMM 5-layer + KL penalty & $2.46$ & $1.36$ & $0.71$ \\
DRMM 5-layer + KL and NN penalties  & $\bf0.91$ & $\bf0.57$ & $\bf0.6$ \\
\hline
\end{tabular}
\vspace{1mm}
\caption{Test error for semi-supervised learning on MNIST using $N_{U} = 60K$ unlabeled images and $N_{L} \in \{100,600,1K\}$ labeled images.}
\label{tab:test-error-mnist}
\end{table*}

\begin{table*}[h]
  \small
  \centering
  \renewcommand{\arraystretch}{1.15}
  \begin{tabular}{cccc}
    \hline
     Model & \multicolumn{3}{c}{Test error reduction (\%)}\\
          & $N_L=50$ & $N_L=100$ & $N_L=1000$ \\
    \hline
    \hline
     DRMM 5-layer + NN penalty & $-0.37$ & $1.13$ & $0.09$\\
     DRMM 5-layer + KL penalty & $19.27$ & $12.05$ & $1.64$\\
     DRMM 5-layer + KL and NN penalties & $20.82$ & $12.84$ & $1.75$\\
    \hline
  \end{tabular}
  \caption{The reduction in test error for semi-supervised learning on MNIST when using the KL penalty, the NN penalty, and both of them. The trainings are for $N_{U} = 60K$ unlabeled images and $N_{L} \in \{50, 100, 1K\}$ labeled images}
  \label{tab:value-added-analysis}
\end{table*}


\subsection{SVHN}

Like MNIST, the SVHN dataset is used for validating semi-supervised learning methods. SVHN contains 73,257 color images of street-view house number digits. For training, we use a 9-layer DRMM with the feedforward propagation similar to the Conv Large network in \cite{rasmus2015semi}. Other training details are the same as for MNIST. We train our model on $N_{L}=500, 1K, 2K$ and show state-of-the-art results in Table~\ref{tab:test-error-svhn}.
\begin{table*}[h]
\vspace{1mm}
    \centering
    \small
    \renewcommand{\arraystretch}{1.15}
    \begin{tabular}{cccc}
    \hline
     Model & \multicolumn{3}{c}{Test error (\%) for a given number of labeled examples}\\
           & 500 & 1000 & 2000\\
    \hline
    \hline
     DGN~\cite{kingma2014semi} & &  $36.02 \pm 0.10$ & \\
     Virtual Adversarial~\cite{miyato2015distributional} & & $24.63$ & \\
     Auxiliary Deep Generative Model \cite{maaloe2016auxiliary} & & $22.86$ & \\
     Skip Deep Generative Model \cite{maaloe2016auxiliary} & & $16.61 \pm 0.24$ & \\
     ImprovedGAN \cite{salimans2016improved}& $18.44 \pm 4.8$ & $8.11 \pm 1.3$ & $\bf6.16 \pm 0.58$ \\
    \hline
    DRMM 9-layer + KL penalty & $11.11$ & $9.75$ & $8.44$ \\
    DRMM 9-layer + KL and NN penalty & $\bf9.85$ & $\bf6.78$ & $6.50$ \\
    \hline
    \end{tabular}
\vspace{1mm}
\caption{Test error for semi-supervised learning on SVHN using $N_{U} = 73,257$ unlabeled images and $N_{L} \in \{500, 1K, 2K\}$ labeled images.}
\label{tab:test-error-svhn}
\end{table*}
		
\subsection{CIFAR10}

We use CIFAR10 to test the semi-supervised learning performance of the DRMM on natural images. For CIFAR10 training, we use the same 9-layer DRMM as for SVHN. Stochastic gradient descent with exponentially-decayed learning rate is still used to train the model.
Table \ref{tab:test-error-cifar} presents the semi-supervised learning results for the 9-layer DRMM for $N_{L}=4K, 8K$ images. Even though we only use a simple SGD algorithm to train our model, the DRMM achieves comparable results to state-of-the-art methods ($21.8\%$ versus $20.40\%$ test error when $N_{L=4K}$ as with the Ladder Networks). For semi-supervised learning tasks on CIFAR10, the Improved GAN has the best classification error ($18.63\%$ and $17.72\%$ test errors when $N_{L}\in\{4K, 8K\}$). However, unlike the Ladder Networks and the DRMM, GAN-based architectures have an entirely different objective function, approximating the Nash equilibrium of a two-layer minimax game, and therefore, are not directly comparable to our model.

\begin{table*}[h]
  \scriptsize
  \centering
  \small
  \renewcommand{\arraystretch}{1.15}
  \begin{tabular}{ccc}
    \hline
     Model & \multicolumn{2}{c}{Test error (\%)}\\
          & 4000 & 8000 \\
    \hline
    \hline
     Ladder network~\cite{rasmus2015semi} & $20.40 \pm 0.47$ & -\\
     CatGAN \cite{springenberg2015unsupervised} & $19.58 \pm 0.46$ & - \\
     ImprovedGAN \cite{salimans2016improved}  & $\bf18.63 \pm 2.32 $ & $17.72 \pm 1.82$\\
    \hline
     DRMM 9-layer + KL penalty & $23.24$ & $20.95$\\
     DRMM 9-layer + KL and NN penalty & $21.50$ & $\bf17.16$\\
    \hline
  \end{tabular}
  \caption{Test error for semi-supervised learning on CIFAR10 using $N_{U} = 50K$ unlabeled images and $N_{L} \in \{4K, 8K\}$ labeled images.}
  \label{tab:test-error-cifar}
\end{table*}

\subsection{Analyzing the DRMM using Synthetic Imagery}

In order to better understand what the DRMM learns during training and how latent variations are encoded in the DRMM, we train DRMMs on our synthetic dataset which has labels for important latent variations in the data and analyze the trained model using linear decoding analysis. We show that the DRMM disentangles latent variations over multiple layers and compare the results with traditional DCNs.   



\paragraph{Dataset and Training:} 

The DRMM captures latent variations in the data \cite{patel2015probabilistic, patel2016probabilistic}. Given that the DRMM yields very good semi-supervised learning performance on classification tasks, we would like to gain more insight into how a trained DRMM stores knowledge of latent variations in the data. To do such analysis requires the labels for the latent variations in the data. However, popular benchmarks such as MNIST, SVHN, and CIFAR10 do not include that information. In order to overcome this difficulty, we have developed a Python API for Blender, an open-source computer graphics rendering software, that allows us to not only generate images but also to have access to the values of the latent variables used to generate the image.

\begin{figure}[t!]
	\centering
	\includegraphics[width=1.0\textwidth]{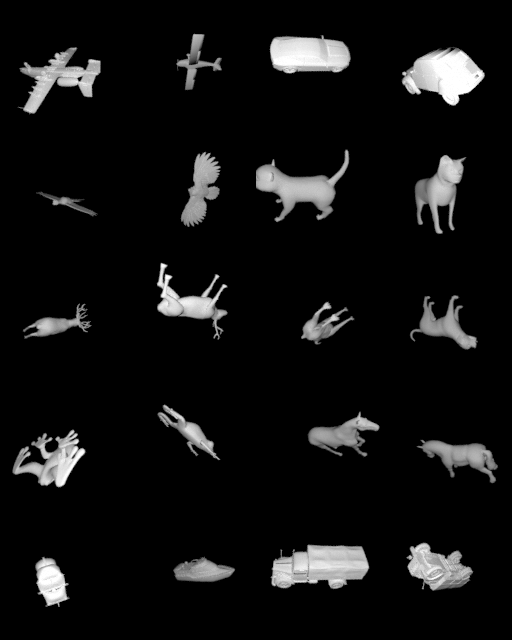}
	\caption{Samples from the synthetic dataset used in linear decoding analysis on DRMM and DCNs}
\label{fig:syn-img}
\end{figure}

The dataset we generate for the linear decoding analysis in Section \ref{sec:linear-decoding} mimics CIFAR10. The dataset contains 60K gray-scale images in 10 classes of natural objects. Each image is of size 32-by-32, and the classes are the same as in CIFAR10. For each image, we also have labels for the slant, tilt, x-location, y-location and depth of the object in the image. Sample images from the dataset are given in Figure \ref{fig:syn-img}.

For the training, we split the dataset into the training and test set, each contains 50K and 10K images, respectively. We perform semi-supervised learning with $N_{L}\in\{4K, 50K\}$ labeled images and $N_{U}=50K$ images without labels using a 9-layer DRMM with the same configuration as in the experiments with CIFAR10. We train the equivalent DCN on the same dataset in a supervised setup using the same number of labeled images. The test errors are reported in Table \ref{tab:test-error-blender}.

\begin{table}[h]
  \small
  \centering
  \renewcommand{\arraystretch}{1.15}
  \begin{tabular}{ccc}
    \hline
     Model & \multicolumn{2}{c}{Test error (\%)}\\
          & $N_L=4K$ & $N_L=$50K \\
    \hline
    \hline
    Conv Large 9-layer & $23.44$ & $2.63$\\
    \hline
     DRMM 9-layer + KL and NN penalty & $\bf6.48$ & $\bf2.22$\\
    \hline
  \end{tabular}
  \caption{Test error for training on the synthetic dataset using $N_{U} = 50K$ unlabeled images and $N_{L} \in \{4K, 50K\}$ labeled images.}
  \label{tab:test-error-blender}
\end{table}

\paragraph{Linear Decoding Analysis:}\label{sec:linear-decoding} 

\begin{figure}
	\centering
	\includegraphics[width=1.0\textwidth]{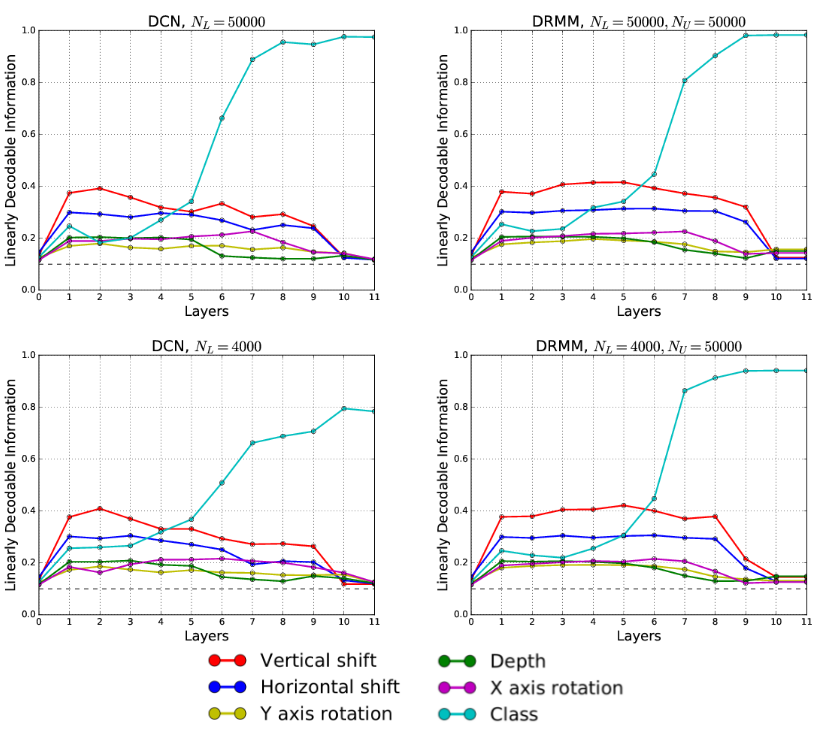}
	\caption{Linear decoding analysis using different numbers of labeled data $N_{L}$. The horizontal dashed line represents  random chance.}
\label{fig:LDI}
\end{figure}

We applied a linear decoding analysis on the DRMMs and the DCNs trained on the synthetic dataset using $N_{L}\in\{4K, 50K\}$. Particularly, for a given image, we map its activations at each layer to the latents variables by first quantizing the values of latent variables into 10 bins and then classifying the activations into each bin using first ten principle components of the activations. We show the classification errors in Figure \ref{fig:LDI}.

Like the DCNs, the DRMMs disentangle latent variations in the data. However, the DRMMs keeps the information about the latent variations across most of the layers in the model and only drop those information when making decision on the class labels. This behavior of the DRMMs is because during semi-supervised learning, in addition to object classification tasks, the DRMMs also need to minimize the reconstruction error, and the knowledge of the latent variation in the input images is needed for this second task.

\section{Conclusions}

In this paper, we proposed a new approach for semi-supervised learning with DCNs. Our algorithm builds upon the DRMM, a recently developed probabilistic generative model underlying DCNs. We employed the EM algorithm to develop the bottom-up and top-down inference in DRMM. We also apply variational inference and utilize the non-negativity constraint in the DRMM to derive two new penalty terms, the KL and NN penalties, for the training objective function. Our method achieves state-of-the-art results in semi-supervised learning tasks on MNIST and SVHN and yields comparable results to state-of-the-art methods on CIFAR10. We analyzed the trained DRMM using our synthetic dataset and showed how latent variations were disentangled across layers in the DRMM. Taken together, our semi-supervised learning algorithm for the DRMM is promising for wide range of applications in which labels are hard to obtain, as well as for future research.


{\small
\bibliographystyle{ieee}
\bibliography{egbib}
}

\clearpage
\appendix





\def\cvprPaperID{2479} 
\def\httilde{\mbox{\tt\raisebox{-.5ex}{\symbol{126}}}}

\ifcvprfinal\pagestyle{empty}\fi




Paper ID 2479
\section{Model Architectures and Training Details}
\begin{table*}[t!]
 \centering
 \small
 \begin{tabular}{lllll}
   \toprule
   Model   & Optimiser/Hyper-parameters & Dataset & \multicolumn{2}{c}{DRMM architecture}    \\
   \midrule
   DRMM  & SGD \cite{lecun1998efficient} & MNIST & Input & 784 (flattened 28x28x1).   \\
   of ConvSmall & learning rate 0.2 - 0.0001 & & E-step Bottom-Up &  Conv 32x5x5 (Full),\\
   \cite{rasmus2015semi} & & & & Maxpool 2x2, \\
   &  over 500 epochs & & & Conv 64x3x3 (Valid), 64x3x3 (Full),\\
   & & & & Maxpool 2x2, \\
   & $\alpha_{H}=1$, $\alpha_{RC}=0.2$, & & & Conv 128x3x3 (Valid), 10x1x1 (Valid),\\
   & & & & Meanpool 6x6, \\
   & $\alpha_{KL}=1$, $\alpha_{NN}=1$ & & & Softmax. \\
   & batch size $= 100$ & & & BatchNorm after each Conv layer. ReLU activation. \\
   & & & Classes & 10. \\
   & & & E-step Top-Down & DRMM Top-Down Reconstruction. No BatchNorm. \\
   & & & & Upsampling nearest-neighbor.\\
   \midrule
   DRMM  & SGD \cite{lecun1998efficient} & SVHN & Input & 3072 (flattened 32x32x3).   \\
    of ConvLarge & learning rate 0.2 - 0.0001 & CIFAR10 & E-step Bottom-Up &  Conv 96x3x3 (Half), 96x3x3 (Full), 96x3x3 (Full),\\
    \cite{rasmus2015semi} & & & & Maxpool 2x2, \\
   &  over 500 epochs & & & Conv 192x3x3 (Valid), 192x3x3 (Full), 192x3x3 (Valid), \\
   & & & & Maxpool 2x2, \\
   & $\alpha_{H}=1$, $\alpha_{RC}=0.5$, & & & Conv 192x3x3 (Valid), 192x1x1 (Valid), 10x1x1 (Valid),\\ 
   & & & & Meanpool 6x6, \\
   & $\alpha_{KL}=0.2$, $\alpha_{NN}=0.5$ & & & Softmax. \\
   & batch size $= 100$ & & & BatchNorm after each Conv layer. ReLU activation. \\
   & & & Classes & 10. \\
   & & & E-step Top-Down & DRMM Top-Down Reconstruction. No BatchNorm. \\
   & & & & Upsampling nearest-neighbor.\\
   \bottomrule
 \end{tabular}
   \caption{Details of the model architectures and trainings in the paper.}
   \label{tab:model_arch}
\end{table*}

The details of model architectures and trainings in this paper are provided in Table~\ref{tab:model_arch}. The models are trained using Stochastic Gradient Descent \cite{lecun1998efficient} with exponentially-decayed learning rate. All convolutions are of stride one, and poolings are non-overlapping. Full, half, and valid convolutions follow the standards in Theano. Full convolution increases the image size, half convolution reserves the image size, and valid convolution decreases the image size. The mean and variance in batch normalizations \cite{ioffe2015batch} are kept track during the training using exponential moving average and used in testing and validation. The implementation of the DRMM generation process can be found in Section~\ref{sec:gen-process}. The set of labeled images is replicated until its size is the same as the size of the unlabeled set (60K for MNIST, 73,257 for SVHN, and 50K for CIFAR10). In each training iteration, the same amounts of the labeled and unlabeled images (half of the batch size) are sent into the DRMM. The batch size used is 100. The values of hyper-parameters provided in Table~\ref{tab:model_arch} are for $N_{L}=100$ in case of MNIST, $N_{L}=1000$ in case of SVHN, and $N_{L}=4000$ in case of CIFAR10. $N_{L}$ is the number of labeled images used in training. Also, DRMM of ConvSmall is the DRMM whose E-step Bottom-Up is similar to the ConvSmall network \cite{rasmus2015semi}, and DRMM of ConvLarge is the DRMM whose E-step Bottom-Up is similar to the ConvLarge network \cite{rasmus2015semi}. Note that we only apply batch normalization after the convolutions, but not after the pooling layers.

\section{Generation Process in the Deep Rendering Mixture Model}
\label{sec:gen-process}
As mentioned in Section~\ref{sec:drmm} of the paper, generation in the DRMM takes the form:
\begin{align}
    c^{(L)} &\sim \textrm{Cat}(\{\pi_{c^{(L)}}\}) \nonumber \\
    g^{(\ell)} &\sim \textrm{Cat}(\{\pi_{g^{(\ell)}}\})\nonumber\\
    \mu_{cg} &\equiv \Lambda_{g}\mu_{c^{(L)}}\nonumber\\
             &\equiv \Lambda^{(1)}_{g^{(1)}}\Lambda^{(2)}_{g^{(2)}} \dots \Lambda^{(L-1)}_{g^{(L-1)}}\Lambda^{(L)}_{g^{(L)}} \mu_{c^{(L)}} \nonumber 
             \\
    I &\sim \N(\mu_{cg},\,\sigma^{2}\textbf{1}_{D^{(0)}}) \nonumber,
    \label{eq:gen3}
\end{align}
where $\ell\in [L] \equiv \{1,2,\dots,L\}$ is the layer, $c^{(L)}$ is the object category, $g^{(\ell)}$ are the latent (nuisance) variables at layer $\ell$, and $\Lambda_{g^{(\ell)}}^{(\ell)}\in \R^{D^{(\ell)} \times D^{(\ell + 1)}}$ are parameter dictionaries that contain templates at layer $\ell$. The image $I$ is generated by adding isotropic Gaussian noise to a multiscale ``rendered'' template $\mu_{cg}$. When applying the hard Expectation-Maximization (EM) algorithm, we take the zero-noise limit. Here, $g^{(\ell)}=\left(t^{(\ell)},a^{(\ell)}\right)$ where $a^{(\ell)}\equiv \left(a_{x^{(\ell+1)}}^{(\ell)}\right)_{x^{(\ell+1)}\in \mathcal{X}^{(\ell+1)}} \in \R^{D^{(\ell+1)}}$ is a vector of binary switching variables that select the templates to render and $t^{(\ell)}\equiv \left(t_{x^{(\ell+1)}}^{(\ell)}\right)_{x^{(\ell+1)}\in \mathcal{X}^{(\ell+1)}}\in \R^{D^{(\ell+1)}}$ is the vector of rendering positions. Note that $x^{(\ell+1)} \in \mathcal{X}^{(\ell + 1)} \equiv \{\textrm{pixels in level } \ell+1\}$ (see Figure~\ref{fig:deep-sparse-coding}B) and $t_{x^{(\ell+1)}}^{(\ell)} \in \{\textrm{UL}, \textrm{UR}, \textrm{LL}, \textrm{LR}\}$ where UL, UR, LL and LR stand for upper left, upper right, lower left and lower right positions, respectively.  As defined in \cite{patel2016probabilistic}, the intermediate rendered image $z^{(\ell)}$ is given by:

\begin{align}
    z^{(\ell)} &\equiv \Lambda^{(\ell)}_{g^{(\ell)}}z^{(\ell+1)} \\
    &= \Lambda^{(\ell)}_{t^{(\ell)},a^{(\ell)}}z^{(\ell+1)} \\
&= \sum_{x^{(\ell+1)}\in \mathcal{X}^{(\ell + 1)}} T^{(\ell)}_{t^{(\ell)}_{x^{(\ell+1)}}} Z^{(\ell)}_{x^{(\ell+1)}}\left(\Gamma^{(\ell)}M^{(\ell)}_{a^{(\ell)}}\right)_{x^{(\ell+1)}}z^{(\ell+1)}_{x^{(\ell+1)}}\\
&= \textrm{DRMMLayer}(z^{(\ell+1)}, t^{(\ell)}, a^{(\ell)}, \Gamma^{(\ell)})
\end{align}
where  $M^{(\ell)}_{a^{(\ell)}}\equiv \textrm{diag}\left(a^{(\ell)}\right) \in \R^{D^{(\ell + 1)} \times D^{(\ell + 1)}}$ is a masking matrix, $\Gamma^{(\ell)}\in \R^{F^{(\ell)} \times D^{(\ell + 1)}}$ is the set of core templates of size $F^{(\ell)}$ (without any zero-padding and translation) at layer $\ell$, $Z^{(\ell)}\in \R^{D^{(\ell)} \times F^{(\ell)}}$ is a set of zero-padding operators, and $T^{(\ell)}_{t^{(\ell)}}\in \R^{D^{(\ell)} \times D^{(\ell)}}$ is a set of translation operators to position $t^{(\ell)}$. Elements of $Z^{(\ell)}$ and $T^{(\ell)}_{t^{(\ell)}}$ are indexed by $x^{(\ell+1)}$. Also, $\Gamma^{(\ell)}[:,x^{(\ell+1)}]$ are the same for $x^{(\ell+1)}$ in the same channel of the intermediate rendered image $z^{(\ell + 1)}$. Note that in the main paper, we call $z^{(\ell)}$ and $z^{(\ell + 1)}$ intermediate rendered templates.

The DRMM layer can be implemented using convolutions of filters $\Gamma^{(\ell)}$, or equivalently, deconvolutions of filters $\Gamma^{(\ell)T}$. $a^{(\ell)}$ and $t^{(\ell)}$ are used to select rendering templates and positions to render, respectively. In the E-step Top-Down Reconstruction, $\hat{t}^{(\ell)}$ and $\hat{a}^{(\ell)}$ estimated in the E-step Bottom-Up are used instead.

\newpage
{\small
\bibliographystyle{ieee}
\bibliography{egbib}
}

\end{document}